\documentclass[%
 reprint,
 amsmath,amssymb,
 superscriptaddress,
 aps,
]{revtex4-2}

\usepackage{graphicx}
\usepackage{dcolumn}
\usepackage{bm}

\usepackage{mathrsfs}
\usepackage[utf8]{inputenc}
\usepackage[T1]{fontenc}
\usepackage{moreverb,url}
\usepackage{multirow}
\usepackage{booktabs}
\usepackage{gensymb}
\usepackage{graphicx}
\usepackage{mathtools}

\usepackage[colorlinks,bookmarksopen,bookmarksnumbered,citecolor=red,urlcolor=red]{hyperref}
\usepackage[dvipsnames]{xcolor}
\definecolor{orange}{rgb}{1 0.5 0}
\definecolor{pink}{RGB}{255 20 147}
\definecolor{darkGreen}{RGB}{0 153 0}

\renewcommand{\eqref}[1]{\textup{{\normalfont Eq.~(\ref{#1}}\normalfont)}}

\begin{document}

\author{Carine Rognon}
    \affiliation{
	Laboratory of Intelligent Systems, École Polytechnique Fédérale de Lausanne (EPFL), Lausanne, Switzerland
    }
\author{Loic Grossen\thanks{C. Rognon and L. Grossen contributed equally to this work.}}
    \affiliation{
	Laboratory of Intelligent Systems, École Polytechnique Fédérale de Lausanne (EPFL), Lausanne, Switzerland
    }

\author{Stefano Mintchev}
    \affiliation{
	Environmental Robotics Laboratory,	Dep. of Environmental Systems Science, ETHZ, Zurich, Switzerland
	\&\\
	Swiss Federal Institute for Forest Snow, and Landscape Research (WSL), Birmensdorf, Switzerland\\
    }

\author{Jenifer Miehlbradt}
    \affiliation{
	Translational Neural Engineering Laboratory, École Polytechnique Fédérale de Lausanne (EPFL), Lausanne, Switzerland
    }

\author{Silvestro Micera}
    \affiliation{
	Translational Neural Engineering Laboratory, École Polytechnique Fédérale de Lausanne (EPFL), Lausanne, Switzerland
    }

\author{Dario Floreano}
 \email{dario.floreano@epfl.ch}
    \affiliation{
	Laboratory of Intelligent Systems, École Polytechnique Fédérale de Lausanne (EPFL), Lausanne, Switzerland
    }


\title{A Portable and Passive Gravity Compensation Arm Support for Drone Teleoperation}

\begin{abstract}
Gesture-based interfaces are often used to achieve a more natural and intuitive teleoperation of robots. Yet, sometimes, gesture control requires postures or movements that cause significant fatigue to the user. In a previous user study, we demonstrated that na\"{\i}ve users can control a fixed-wing drone with torso movements while their arms are spread out. However, this posture induced significant arm fatigue. In this work, we present a passive arm support that compensates the arm weight with a mean torque error smaller than 0.005 $\frac{\text{N}}{\text{kg}}$ for more than 97$\%$ of the range of motion used by subjects to fly, therefore reducing muscular fatigue in the shoulder of on average 58\%. In addition, this arm support is designed to fit users from the body dimension of the 1$^{st}$ percentile female to the 99$^{th}$ percentile male. The performance analysis of the arm support is described with a mechanical model and its implementation is validated with both a mechanical characterization and a user study, which measures the flight performance, the shoulder muscle activity and the user acceptance.
\end{abstract}


\maketitle

\section{Introduction}

\begin{figure*}[tb]
    \centering
    \includegraphics[width=2\columnwidth]{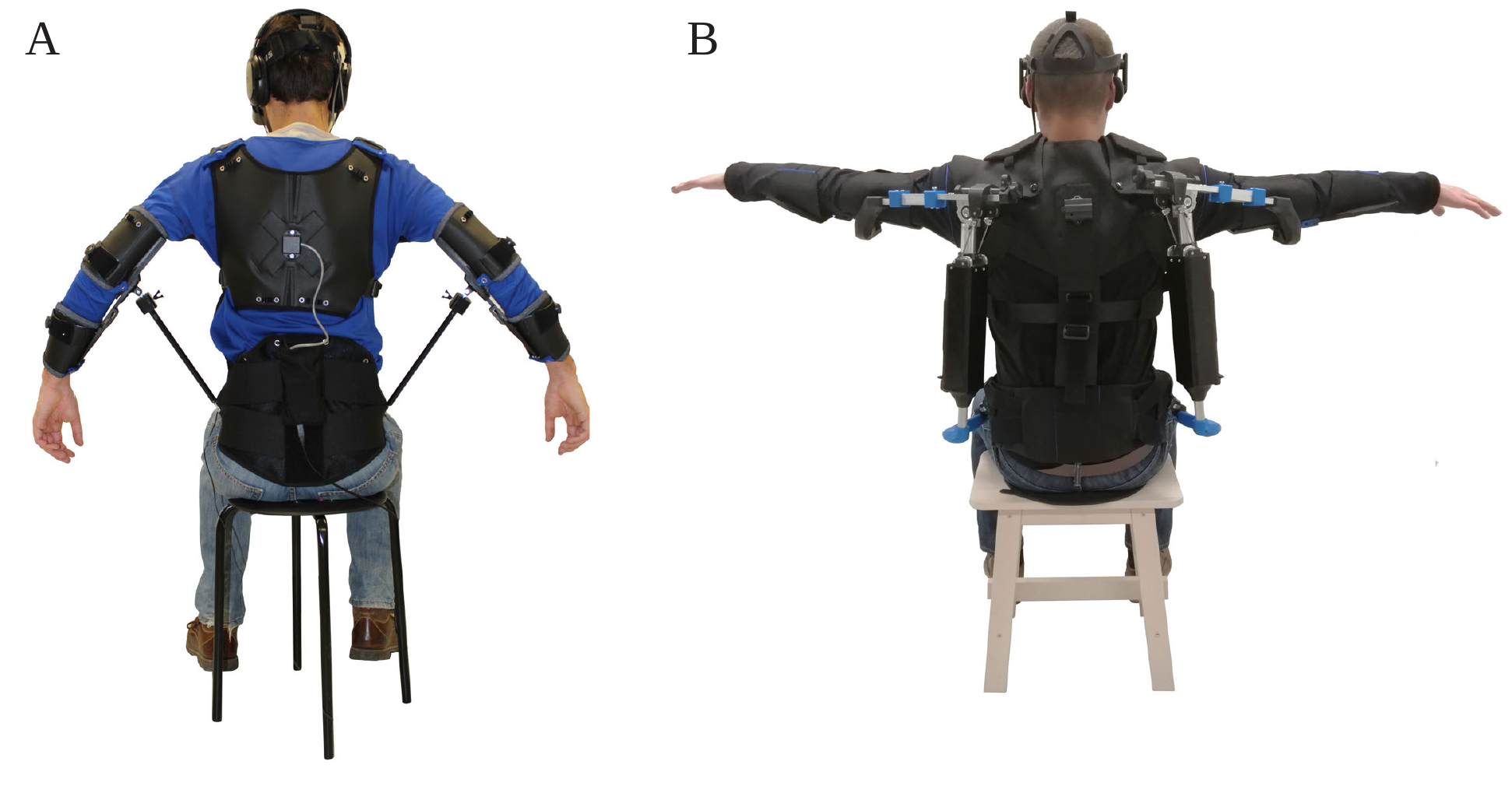}
    \caption{A) The former version of the arm support, called Gas Spring Support (GSS) \cite{rognon2018flyjacket} B) The novel arm support presented in this paper, called Static Balancing Support (SBS)}
    \label{fig:GSS}
\end{figure*}
Teleoperation allows combining the resilience, precision, and force of robots with the cognition of human operators. It requires user interfaces that record the operator's commands and translate them to control inputs for the robot. However, current interfaces, such as joysticks and remote controllers, are often complicated to handle since they require a cognitive effort and a certain level of practice. Gesture-based control of robots can simplify the interaction allowing more intuitive teleoperation~\cite{miehlbradt2018data, sanna2013kinect, pfeil2013exploring, ng2011collocated}. Indeed, gesture control can promote natural and intuitive interactions, as the users' gestures are directly mapped to control inputs for the robot, and there is no need to learn to handle an external object, such as a remote controller.

Yet, sometimes, intuitive and immersive gesture control forces users to take postures or execute movements that can induce a significant muscular fatigue. An example is the FlyJacket, a portable exoskeleton which enables drone teleoperation with intuitive upper body movements \cite{rognon2018flyjacket}. With the FlyJacket, users control the attitude of a drone by bending the torso forward and backward in the sagittal plane, while steering is achieved with a combination of bending on the sides in the frontal plane and twisting in the transverse plane \cite{miehlbradt2018data}. 
A user study demonstrated that the participants' performance increased in a posture with the arms spread out, even if they were not directly used to control the drone \cite{rognon2018flyjacket}. However, after few minutes, they endured arm fatigue and lowered their arms, which influenced their performance.

To prevent this behavior, an arm support needed to be embedded in the portable exoskeleton. 
Portable arm supports in the form of exoskeletons have been developed for human power enhancement
\cite{kobayashi2007development} or to help workers in their tasks \cite{stadler2014robo, stadler2017robo,constantinescu2016optimisation}. Ekso Bionics (Richmond, CA, USA) commercialized an upper body exoskeleton for fatigue reduction during overhead manufacturing, assembly, and construction. 
Research has also focused on exoskeletons for shoulder support in the frame of rehabilitation~\cite{roderick2005design, sugar2007design}. These devices have the advantage to be portable, enabling the patient to perform daily life activities at home. To enhance user acceptance, compliance to the body, and safety, soft exoskeletons have been in the focus of recent work \cite{park2017development, o2017soft, simpson2017exomuscle, cappello2016design, lessard2018soft}. Pneumatically actuated arm supports allow a safe interaction with the human body due to their intrinsic compliance. Yet, to be actuated they require a compressor, which limits the portability. Soft supports (i.e. pneumatic or cable-driven exoskeletons) are compliant and usually more lightweight than rigid structures but their efficiency and force transmission are significantly lower.

For the first version of the FlyJacket, we developed a straightforward passive arm support, called Gas Spring Support (GSS) (see \cite{rognon2018flyjacket} and Figure \ref{fig:GSS}). It is composed of a gas spring placed between the waist and the upper arm of the user. With this basic arm support, the users could rest their arms on the gas springs and the sensation of fatigue was significantly reduced.
However, the GSS presents three main limitations that we want to address with the new arm support proposed in this paper. First, the GSS limits the Range of Motion (ROM) of the arm. Even if the GSS was designed to allow some movements that are required outside the flight task, such as manipulating virtual reality goggles, the restricted ROM hampers other manipulations such as using a computer or reaching objects on a shelf. Second, the weight compensation provided by the GSS is limited to a very narrow region of the ROM. When using the GSS, the arm weight is fully compensated only at one position when the force of the gas spring is equal to the arm weight. At other positions, the arm weight is only partially compensated by the force relative to the compression of the inner gas and this compensation varies with the position. Therefore, some arm positions can still induce fatigue when they are maintained for a relatively long period. Third, the GSS has a limited adaptability to users' morphology. Springs with different lengths and stiffnesses must be used to support arms with different weights and lengths.

In this article, we present a portable Static Balancing Support (SBS) that can cover more than 97$\%$ of the ROM required to fly with the FlyJacket while being adaptable to 98$\%$ of the population. First, the device's design is detailed and compared to the GSS in a simulation. Then, the performances of the new SBS device are characterized mechanically and with a user study. For the mechanical characterization, the output torque of the SBS was measured and compared with the torque induced by the arm weight. For the user study, twelve participants performed a flight task with three different arm support conditions -- SBS, GSS, and No Support (NS) -- and their performance and shoulder muscle activity were compared.

\section{Analysis of the Requirements}
The FlyJacket is an exosuit conceived to intuitively and naturally fly a drone using a combination of torso gestures \cite{miehlbradt2018data, rognon2018flyjacket}. The use of soft fabrics guarantees comfort during operation and adaptability to different body morphologies. The novel arm support, SBS, is conceived as a separate module that can be added to the existing FlyJacket. The function of the device is to compensate the arm weight while remaining transparent to the user. For this purpose, the arm support should fulfill three main requirements: 
\begin{enumerate}
\item it should balance the weight for every position assumed by the arm during operation,
\item its workspace should be compatible with the user ROM during operation,
\item its structure and balancing strength should adapt to different users’ morphologies, in particular different upper body lengths and arms weights.
\end{enumerate}

The aim of the FlyJacket is to achieve an immersive control of a fixed-wing drone using natural motion. \cite{miehlbradt2018data} determined the most intuitive body gestures to control the drone. As described above, these movements are the following: leaning the torso forward and backward controls the pitch down and up respectively, and a combination of laterally bending and rotating the torso controls the roll of the drone on the corresponding side. In most case, users adopt a position with the arms spread out, which showed to improve their performance.

These gestures are produced at two main locations; the waist and the shoulder (see Figure~\ref{fig:ModelSchema}). Although sagittal bending of the torso uses the hip joint while the lateral bending uses the lumbosacral joints, we approximated that both joints are collocated. The required ROMs of these two joints are based on the data collected directly on users during previous experiments \cite{miehlbradt2018data} and \cite{rognon2018haptic} and are summarized in Table~\ref{tab:rom}. However, not all angle combinations were observed and the participants displayed smaller joint angles most of the time. The zero position ($\alpha = 0\degree$) for the shoulder corresponds to the position in which the arm is horizontally extended, perpendicular to the torso. The ROM corresponding to each body dimension can be calculated with Equations \ref{eq:xROM} and \ref{eq:yROM} (see Appendix). An example of arm ROM is illustrated by the purple region in Figure~\ref{fig:ModelSchema}.

To address the adaptability to different morphologies, the device is modelled based on body dimensions from the 1$^{st}$ percentile female, hereafter denoted as "1PF", until the 99$^{th}$ percentile male, denoted "99PM", size and body mass of the US population~\cite{tilley2002measure} and \cite{clauser1969weight}. Using these data, the device would fit 98 $\%$ of the population. The dimensions for the upper body are summarized in Table~\ref{tab:size}.

\begin{table}[tb]
    \caption{Range of Motion Used During Flight}
    \centering
    \begin{tabular}{|l|c|c|}
        \hline
        \multirow{ 2}{*}{\textbf{Body Movement}} & \multicolumn{2}{|c|}{\textbf{Angle (\degree)}}\\
         & Min & Max \\ \hline
         Torso flexion / extension & -30 & 40 \\ \hline
         Torso lateral bend & -20 & 20 \\ \hline
         Torso rotation & -60 & 60 \\ \hline
         Shoulder abduction & -60 & 5 \\ \hline
         Shoulder flexion & -40 & 0 \\ \hline
         Shoulder rotation & \textit{-5} & \textit{5} \\ \hline 
    \end{tabular}
    \label{tab:rom}
\end{table}

\begin{table*}[tb]
    \caption{Body Dimensions for the 1$^{st}$ Percentile Female and the 99$^{th}$ Percentile Male. See Figure~\ref{fig:ModelAppend} in the Appendix for the location of the dimensions}
    \centering
    \begin{tabular}{|l|c|c|}
        \hline
        \textbf{Body Parts} & \textbf{1$^{st}$ Percentile Female} & \textbf{99$^{th}$ Percentile Male}\\ \hline
         Total arm mass $A_m$ (kg) & 1.86 & 6.56 \\ \hline  
         Upper arm length $L_a$ (m) & 0.234 & 0.312\\ \hline
         Upper trunk height $U_{th}$ (m) & 0.269 & 0.353\\ \hline
         Lower trunk height $L_{th}$ (m) & 0.213 & 0.265\\ \hline
         Half chest width $C_{w}$ (m) & 0.138 & 0.203\\ \hline
         Half hip width $H_w$ (m) & 0.142 & 0.214\\ \hline
    \end{tabular}
    \label{tab:size}
\end{table*}
\section{Design and Working Principle}
The novel arm support is composed of an articulated mechanism that can accompany the user's upper body gestures, and an elastic element to passively balance the arm weight. The mechanism consists of an arm and a torso segment connected with a revolute joint (see Figure \ref{fig:ModelSchema}). The arm segment is fixed on the upper arm, and the torso segment is connected on the hip where the arm weight is redirected. The hip has a higher load tolerance comparing to other parts of the upper body, which makes it a good fixation point to redirect the load \cite{scribano1970design}.

To fulfill the first requirement (1), the arm support resorts to a gravity compensation strategy known as static balancing~\cite{lu2011passive}.
A spring integrated in the torso segment generates a force to balance the arm weight ($W$). The spring force is transmitted to the arm segment by a cable with a lever arm $\Delta s$. The result is a torque ($\Gamma$) that rotates the arm segment upward balancing the arm weight at any position within the workspace of the mechanism. Differently to active balancing exoskeletons with electric or pneumatic motors, static balancing is implemented with passive mechanisms that do not require any source of energy. This solution minimizes the bulkiness of the device and foster portability by decreasing dependency from power sources.

The second requirement (2) is met through the kinematics of the arm support, which are designed to minimize the mismatches between the mechanism workspace and the ROM of the arm. This result is achieved by integrating a revolute joint at the connection between the arm segment and the upper arm, and a spherical joint to connect the torso segment to the hip. The combination of these joints allows to accommodate the majority of the body movements described in Table \ref{tab:rom}.

Requirement 3 is fulfilled by the possibility to adjust the length of the arm and hip segments, and the lever arm $\Delta s$ of the spring to match different user morphologies and arm weights respectively.

\subsection{Performance Analysis}
\begin{figure}[tb] 
    \centering
    \includegraphics[width=\columnwidth]{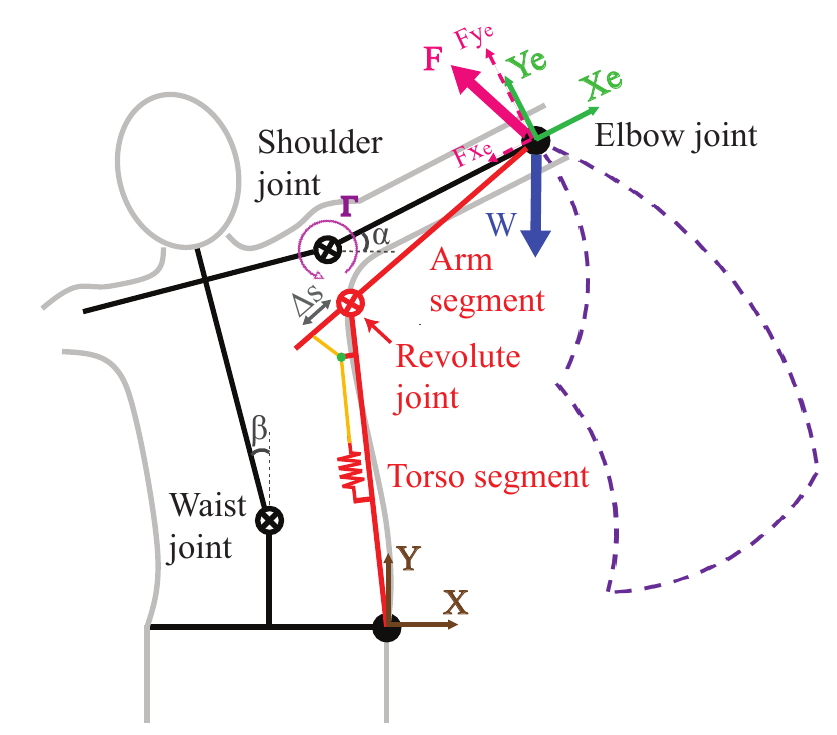}
    \caption{Schema of the Static Balancing Support (SBS) and the induced forces and torque. Two different coordinate systems are used: The $XY$ coordinate system for the positions and the $X_{e}Y_{e}$ coordinate system for the force projections. The force generated by the mechanism can be decomposed in two components in the $X_{e}Y_{e}$ coordinate system: the gravity compensation contribution on $Y_{e}$ and the parasitic forces in the arm on $X_{e}$. Example of ROM domain is shown in purple dashed line.}
\label{fig:ModelSchema}
\end{figure}

The performance of the arm support, in particular its balancing efficacy, undesired parasitic effects, and workspace, have been simulated using MATLAB according to the analytical model presented in the Appendix.  
To evaluate the balancing efficacy and the undesired parasitic effects, it is convenient to consider the balancing force $F$ generated by the arm support at its attachment point near the elbow (see Figure~\ref{fig:ModelSchema}). This force can be projected in the local reference frame $XeYe$ and decomposed into two components: the gravity compensation contribution ($F_{ye}$) that is responsible for the balancing torque $\Gamma$ that limits user fatigue, and a parasitic force ($F_{xe}$) directed along the arm, which produces an undesired sliding motion of the fixation of the arm support along the user's arm. The balancing efficacy can be quantified with the torque error normalized by the arm weight:

\begin{equation}
    T_{errV.norm} = \frac{F_{errY_e} \cdot L_a}{A_m}
    \label{eq:torquerrV}
\end{equation}

\noindent while a measure of the undesired parasitic effect is the force along $X_e$ normalized by the arm weight:

\begin{equation}
    F_{X_e,norm} =  \frac{F_{X_e}}{A_m}
    \label{eq:fpara}
\end{equation}

\noindent with the variables described in Table \ref{tab:size} and in the Appendix.

Figure~\ref{fig:AnaFv} shows the results for the normalized torque error ($T_{errV.norm}$) for both the Static Balancing Support (SBS) (Figure~\ref{fig:AnaFv} A for the 1PF and B for the 99PM) and the Gas Spring Support (GSS) (Figure~\ref{fig:AnaFv} C for the 1PF and D for the 99PM) within the arm ROM during flight. We can observe that with the SBS, almost the entire ROM can be reached (98.2$\%$for the 1PF and 97.8$\%$ for the 99PM) with the exception of the very bottom region. Indeed, the users cannot abduct their arm at -65$\degree$ if they are bending their torso at 20$\degree$ on the same side. However, this range is rarely used by the participants during the flight. On the other hand, when using the GSS, the user can reach only a sub-portion of the ROM required for the flight (58.4$\%$ for the 1PF and 24.4$\%$ for the 99PM). 
For the SBS, the normalized torque error ($T_{errV.norm}$) is relatively uniform and small along the ROM with a mean value of -0.005 $\frac{\text{Nm}}{\text{kg}}$ for the 1PF and -0.002 $\frac{\text{Nm}}{\text{kg}}$ for the 99PM over the ROM and a maximum of -0.024 $\frac{\text{Nm}}{\text{kg}}$ and -0.101 $\frac{\text{Nm}}{\text{kg}}$ respectively, at the lower part of the ROM. For the GSS, the normalized torque error is more important with a mean value more than 130 times higher for the 1PF (0.684 $\frac{\text{Nm}}{\text{kg}}$) and 30 times for the 99PM (-0.058 $\frac{\text{Nm}}{\text{kg}}$) and a maximum of 1.915 $\frac{\text{Nm}}{\text{kg}}$ and -1.411 $\frac{\text{Nm}}{\text{kg}}$ respectively. The normalized torque error of the GSS fluctuates considerably with the arm movement.

\begin{figure}[tb]
    \centering
    \includegraphics[width=\columnwidth]{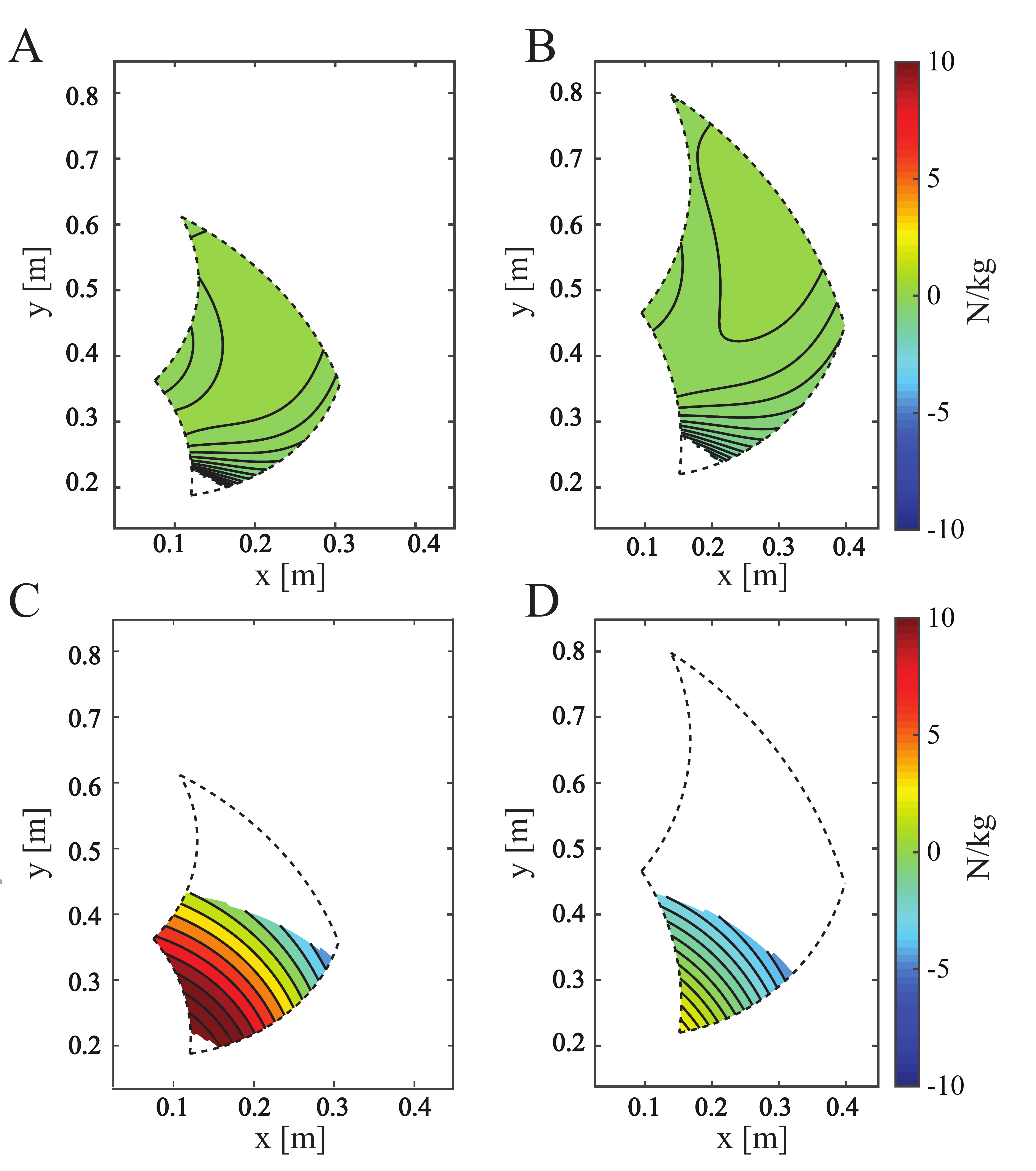}
    \caption{Normalized parasitic force ($F_{X_e,norm}$) over the reachable ROM. A) For the SBS for the 1PF. B) For the SBS for the 99PM. C) For the GSS for the 1PF. D) For the GSS for the 99PM.}
    \label{fig:AnaFu}
\end{figure}

The normalized parasitic force ($F_{X_e,norm}$) is shown in Figure~\ref{fig:AnaFu} A (1PF) and B (99PM) for the SBS, and C (1PF) and D (99PM) for the GSS. The parasitic force is considerably smaller for the SBS with a mean of -0.025 $\frac{\text{N}}{\text{kg}}$ for 1PF and -0.009 $\frac{\text{N}}{\text{kg}}$ for 99PM and a maximum value of -1.026 $\frac{\text{N}}{\text{kg}}$ and -0.325 $\frac{\text{N}}{\text{kg}}$ respectively. On the opposite, the GSS has a parasitic force of 2.922 $\frac{\text{N}}{\text{kg}}$ (1PF) and -0.187 $\frac{\text{N}}{\text{kg}}$ (99PM) with maximum normalized parasitic forces of 8.184 $\frac{\text{N}}{\text{kg}}$ (1PF) and -0.677 $\frac{\text{N}}{\text{kg}}$ (99PM). 

Both the normalized torque error ($T_{errV.norm}$) and the normalized parasitic force ($F_{X_e,norm}$) of the GSS show iso-lines oriented parallel to the torso movement, which signifies that the compensation is not notably influenced by lateral movement of the torso but by the arm angle, which changes significantly both the normalized torque error and the parasitic force.


\begin{figure}[tb]
    \centering
    \includegraphics[width=\columnwidth]{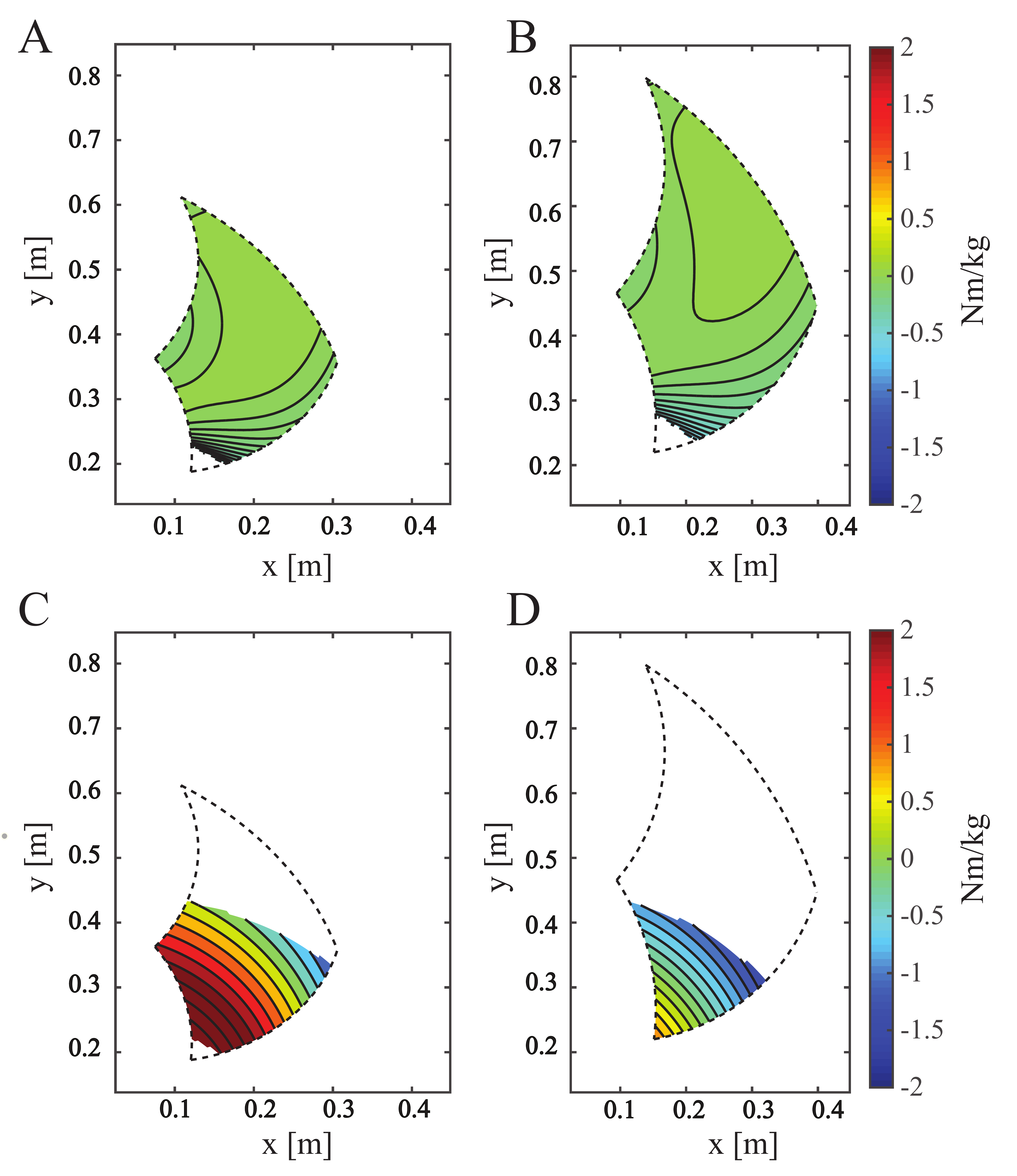}
    \caption{Normalized torque error ($T_{errV.norm}$) over the reachable ROM. A) For the SBS for the 1PF. B) For the SBS for the 99PM C) For the GSS for the 1PF. D) For the GSS for the 99PM.}
    \label{fig:AnaFv}
\end{figure}

\section{Implementation}

\begin{figure}[tb]
    \centering
    \includegraphics[width=\columnwidth]{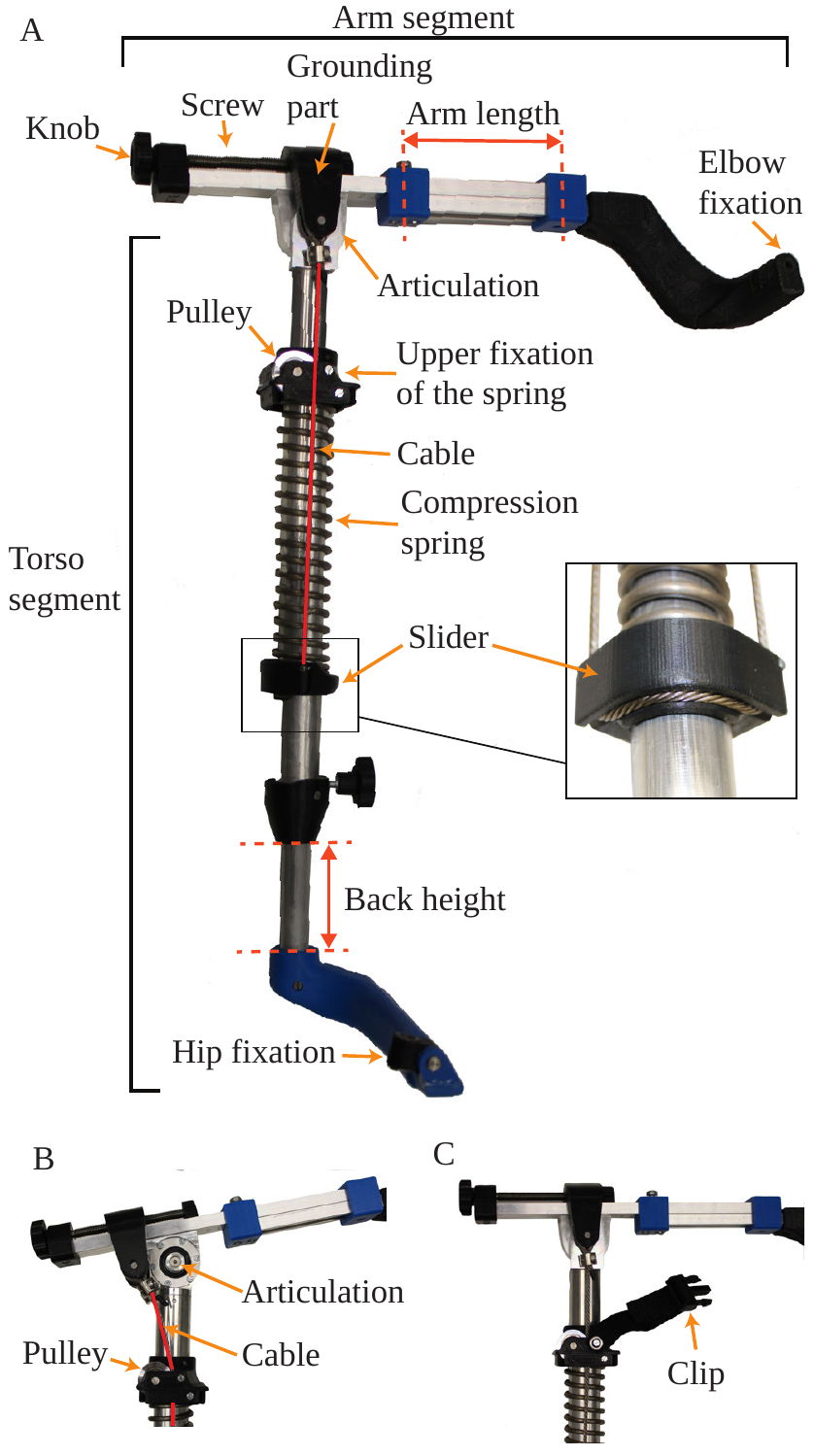}
    \caption{A) Full device with an inset showing the turning point of the cable. B) Grounding part shifted showing the cable angle and the bearing of the articulation. C) Clip to maintain the device close to the torso. Cables are highlighted in red.}
\label{fig:Device}
\end{figure}

The new arm support is shown in Figure~\ref{fig:Device}. 
The articulation linking both the arm and the torso segments is a revolve joint made of a bearing (3200 A-2RS1TN9/MT33, SKF, Sweden) to minimize friction. This articulation is sized to sustain a torque ($\Gamma_p$) of 10 Nm in the sagittal plane, which happens when the maximal arm mass, i.e. 6.56 kg, is at an out of lateral plane distance of 150 mm. This torque corresponds to the offset of the arm regarding the device, i.e. the body and the device are not located on the same lateral plane (see Figure~\ref{fig:userStudy} E). 

A compression spring (Ressorts du Leman Sarl, Switzerland) is generating the balancing force (Figure \ref{fig:Device}). Compression springs were preferred over tension springs as, in most cases, they produce more force for the same dimensions and they do not need to be preloaded. The compression spring is guided by the torso segment to avoid risks of buckling and is fixed at its upper extremity.
The lower extremity of the spring is attached to a slider, which transfers its linear motion to a cable (three millimeters steel, Jakob AG, Switzerland, highlighted in red in Figure~\ref{fig:Device} A and B). The cable is turning in a groove around the slider (see inset of Figure~\ref{fig:Device} A), allowing to balance its tension to each side of the spring. The robustness of this slider was assessed by running a finite element simulations (SolidWorks, Dassault Systèmes, France) to ensure that it could sustain the full load of the spring (up to 600 N). 

At its extremities, the cable is attached to a grounding part. This part can translate along a screw (M10 x 1.5) by rotating a knob located at the extremity of the arm segment (see Figure~\ref{fig:Device}). Thus, the tension in the cable can be changed and the force of the gravity compensation can be adapted to each user. Ten millimeters of the grounding part course corresponds to one kg of arm mass compensation for the largest arm length (0.312 m). When the grounding part is aligned with the articulation as in Figure~\ref{fig:Device} A, the torque compensation of the system is null. The redirection of the cables is done with pulleys mounted on bearings (626-2Z, SKF, Sweden) for less friction. To prevent the unwanted rotation of the grounding part in case of unequal tension in both cables, it is linearly guided by the screw, mounted on plain bearings for less friction, and by the linear profile of the arm segment.  

Two distances (arm length and back height) can be adapted 
to the user's morphology (see location in Figure~\ref{fig:Device} A).
The device is fixed to the hip using a plate attached to the waist belt of the FlyJacket (see Figure~\ref{fig:userStudy} C). As the entire load of the arm is redirected toward this fixation point, its dimensions (0.11 x 0.07 m) were designed to have a large force distribution area, reducing the risk of pain. The link between the device and this hip plate is made through a three degrees of freedom (DOF) joint made from two plain bearings to ensure a low friction in the joint. To remove the arm support from the FlyJacket, it can be disconnected from the hip plate using a pin. The device was linked to the upper arm support using a hinge joint similar to that used in the GSS (see details in \cite{rognon2018flyjacket}). Special care was given to the design of these two fixation points to avoid contact between the device and the body, which would reduce the ROM and may injure the user. An additional attachment point was set to the FlyJacket with a clip between the upper fixation of the spring and the top of the shoulder to maintain the device close to the torso (Figure~\ref{fig:Device} C and Figure~\ref{fig:userStudy} A). 

To avoid any injuries due to the compression spring (such as having body part, hairs, or clothes being pinched), it was covered by a plastic box, which also creates a protection from the system in case of failure (Figure~\ref{fig:userStudy}). When the device is not being used, the spring is still under tension. To avoid any device damage or user injuries, the spring can be locked by an insert placed below the spring on the torso segment to prevent an unwanted loading of the spring (Figure~\ref{fig:userStudy} D). The insert can not be removed if the spring is compressing it.

The mass of the full device is 1.6 kg per arm support and its collapsed size 0.53 x 0.20 x 0.13 m.

\section{Experimental Validation}
The device was validated with both a mechanical characterization and a user study.

\subsection{Mechanical Characterization}
\subsubsection{Experimental Procedure}
\begin{figure}[tb]
    \centering
    \includegraphics[width=\columnwidth]{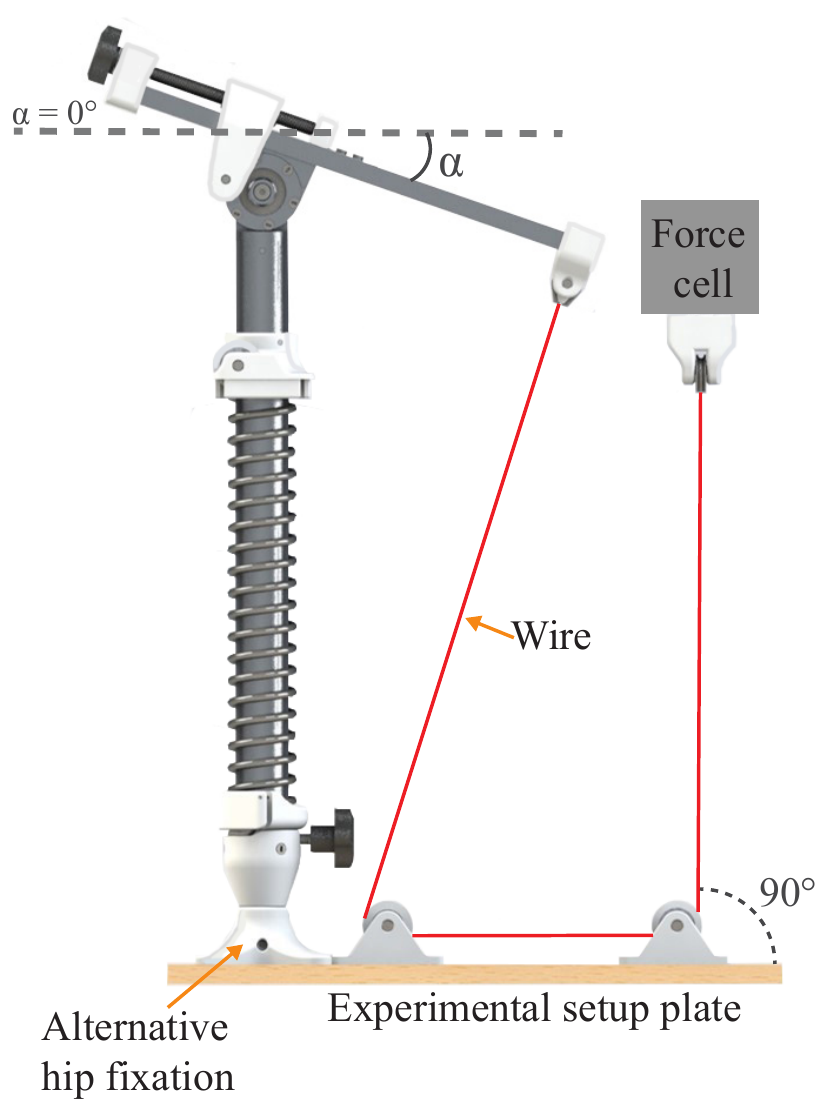}
    \caption{Setup for the mechanical characterization}
    \label{fig:setupmecha}
\end{figure}
The torque response of the device was measured when varying the angle of the arm segment using an INSTRON machine (Figure~\ref{fig:setupmecha}). The device was vertically attached to a rigid plate with an alternative hip fixation and the arm segment was vertically pulled down by a wire (Dyneema 0.4mm, Spiderwire, SC, USA) connected to the force cell of the INSTRON machine.  
The output force was measured for arm segment angles between -69$\degree$ and 63$\degree$ (negative angles corresponding to the arm segment going down). This range was intentionally limited to be smaller than the actual ROM of this segment (-80$\degree$ to 80$\degree$) in order to protect the device against collisions or singularities. In addition, due to the rigid connection between the torso segment and the plate, the system undergoes high vertical forces outside of this range, which may damage the device.

Tests were conducted by changing the grounding part position from 10 mm to 60 mm, with steps of 10 mm, with the arm length set at its largest distance (0.312 m). This corresponds to a weight compensation from one to six kg. The output force for each grounding part position was measured over ten cycles. The corresponding theoretical torque was computed using Equation~\ref{eq:tocalculatetorque} described in the Appendix.

\subsubsection{Results}

\begin{figure}[tb]
    \centering
    \includegraphics[width=\columnwidth]{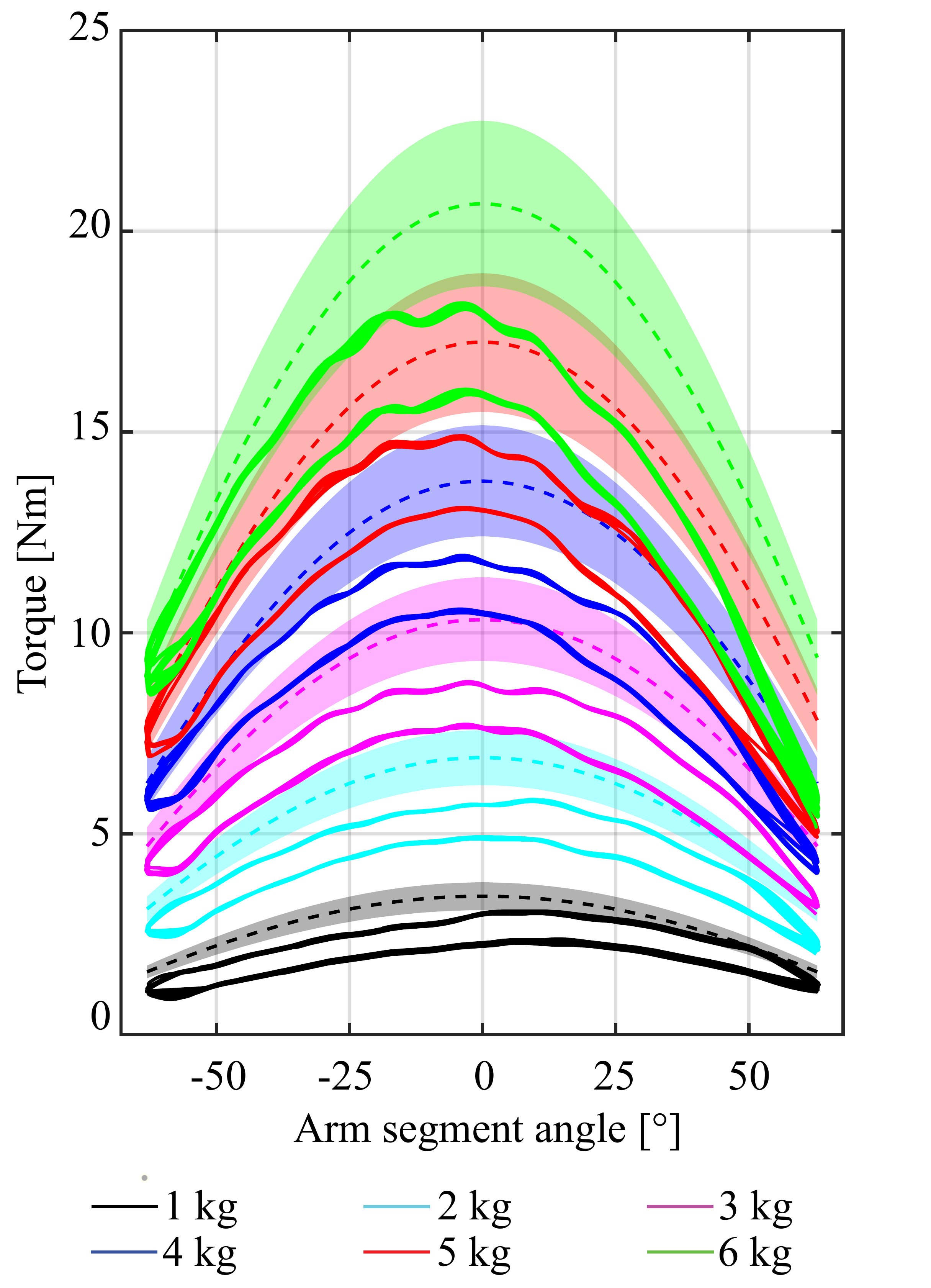}
    \caption{Theoretical torque (dashed lines), theoretical torque uncertainties due to the 10$\%$ tolerance on the spring force constant given by the manufacturer (shaded areas), and measured torque (solid line). The legend (weights between one and six kilos) corresponds to the weight that can be lifted when the trimming is done for the greatest arm distance (0.312 m)}
    \label{fig:torqueRes}
\end{figure}

The torque response over the arm segment angles is shown in Figure~\ref{fig:torqueRes}. Dashed lines represent the theoretical torque, the shaded areas show the theoretical torque uncertainties due to the 10$\%$ tolerance on the spring force constant given by the manufacturer, and the solid lines represent the measured torques (see Appendix for the calculations). 

The measured torques follow the same trend as the theoretical torque; i.e. they have a sinusoidal shape over the ROM of the arm segment and increased proportionally to the arm's weight. However, the measured torques were lower than the theoretical torque. 
A possible reason could be some friction between the slider and the torso segment. This could be solved by adding a linear bearing between the slider and the segment but this would add complexity to the system. On the other hand, despite the fact that the maximal output torque is below expectation, the tuning range is still fairly large and can adapt a variety of arm weights and sizes.
The maximal output torque of 18 Nm still ensures arm masses going up to 5.87 kg for the maximal arm length (0.312 m) to be fully compensated. Indeed, during the user study (see next section) the arm weight of all participants could be compensated. We can also observe a shift of the maximum locations for the measured torque from negative angles for the highest weights to positive angles for the lowest weights. This is due to the change in cable length between the pulley and the grounding part (called $b$ in Figure~\ref{fig:ModelAppend} B), which slightly increases relatively to the displacement of the grounding part (i.e. when $\Delta s$ increases). This length has been considered as constant to simplify the theoretical model. As the average of the population has an arm weight between 3 and 4 kg, a maximum at zero angle has been set for these weights. 

\begin{figure}[tb]
    \centering
    \includegraphics[width=\columnwidth]{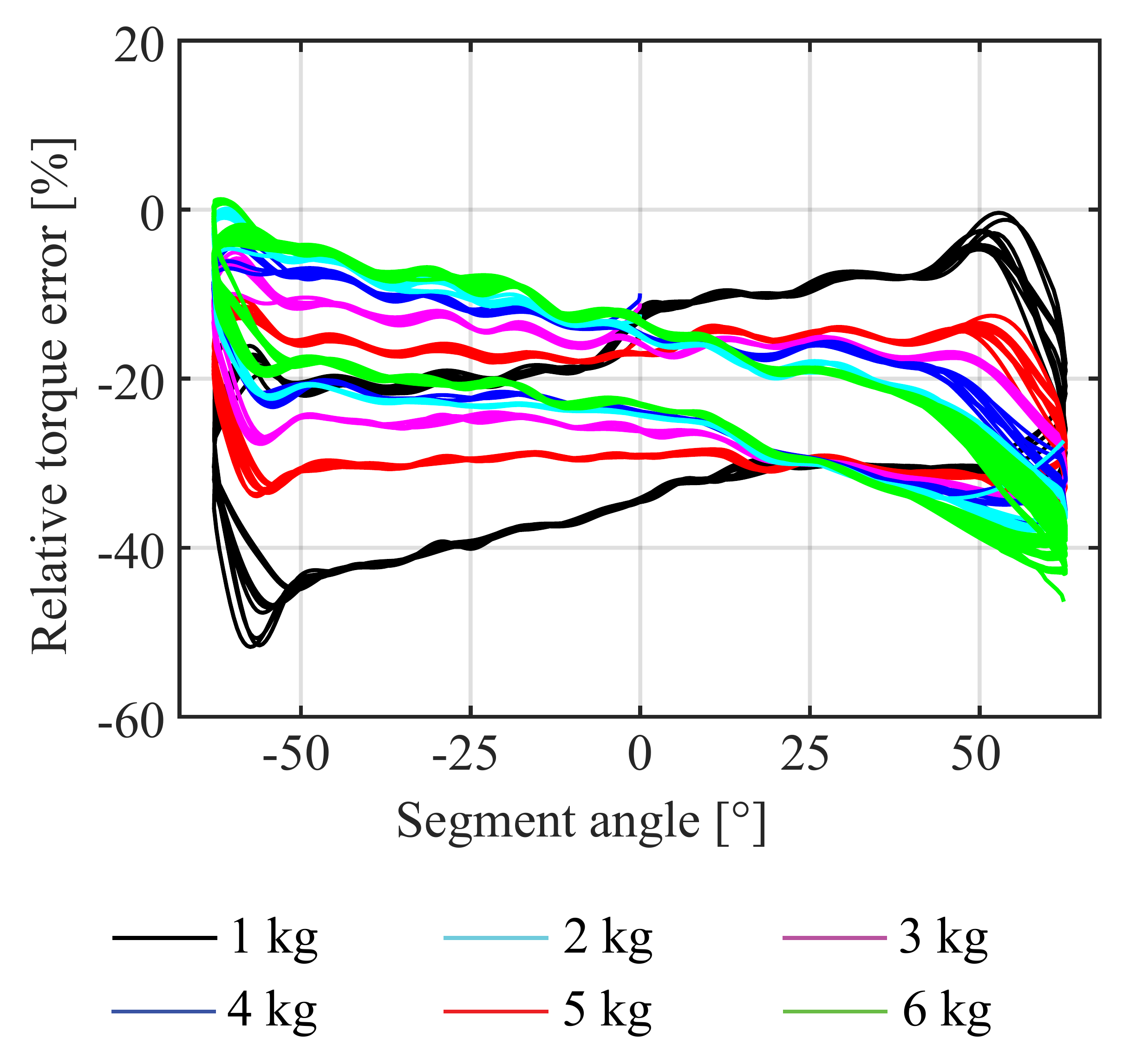}
    \caption{Relative torque error. The legend (weights between one and six kilos) corresponds to the weight that can be lifted when the trimming is done for the greatest arm distance (0.312 m)}
    \label{fig:torqueRelErr}
\end{figure}

Figure~\ref{fig:torqueRes} also shows hysteresis in the torque, which can be due to the friction of the slider against the torso segment or due to the testing setup (i.e. energy loss in the pulleys and bearings, both cable and wire friction and deformation, 3D printed part deformation, etc.). Figure~\ref{fig:torqueRelErr} displays the relative torque error; the difference between the theoretical curve and the measured data. We can observe that the relative torque error and the hysteresis tend to be larger for smaller arm weights (such as the black curve) reaching 50$\%$ for an arm segment angle of -60$\degree$, that is when a small output torque needs to be compensated. This higher relative torque error can be explained because of the larger relative influence of the friction for small torque than for larger torque.

\subsection{User Study}
\label{sec:userstudy}

\subsubsection{Experimental Procedure}

 \begin{figure}[tb]
    \centering
    \includegraphics[width=\columnwidth]{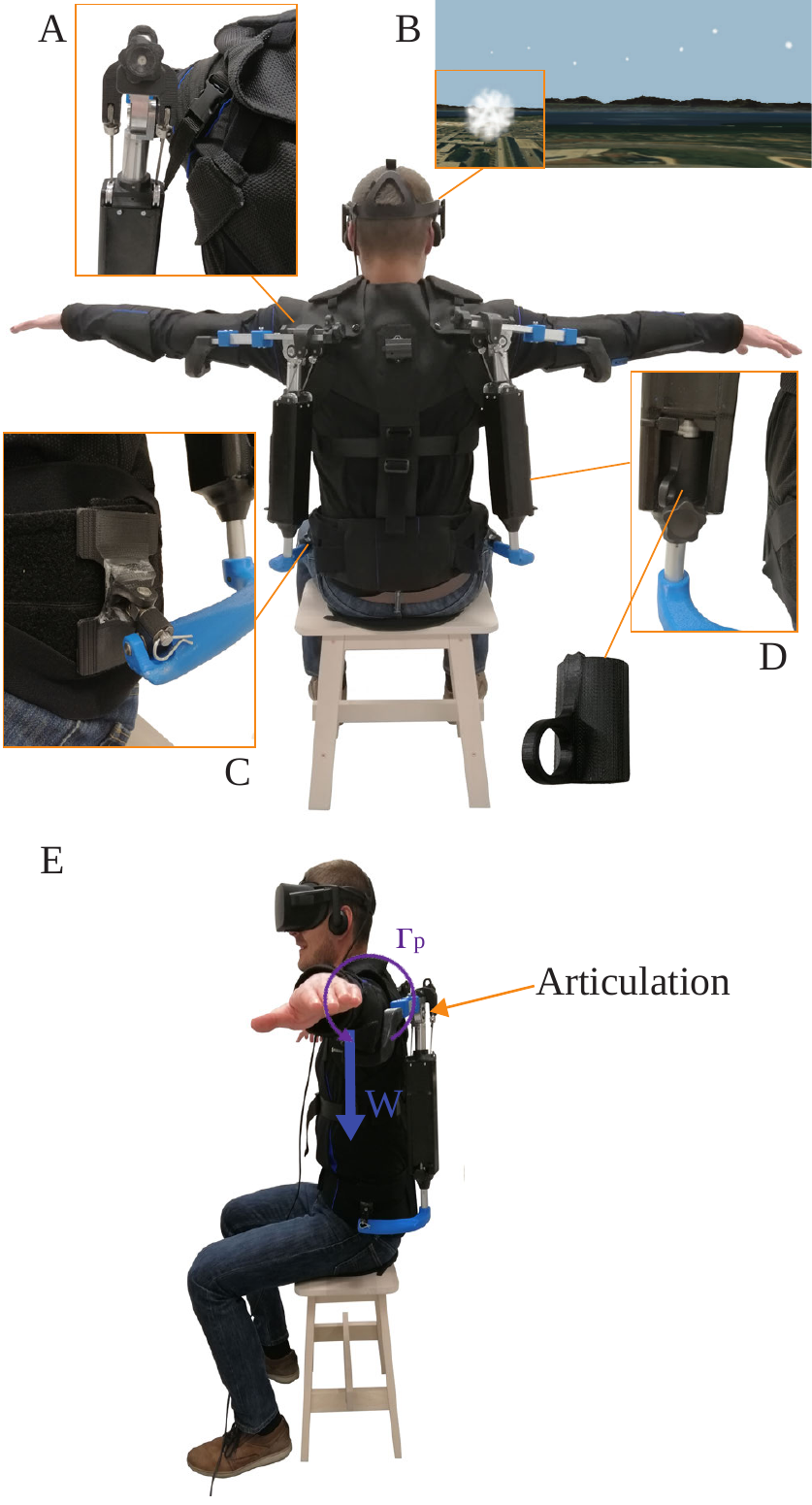}
    \caption{User study setup. A) Insert highlighting the clip to fix the device to the jacket at the shoulder. B) Flight environment with waypoints (symbolized by white clouds) with an insert showing how a waypoint is seen by the participants when they fly towards it. C) Hip fixation point of the device. D) Locker system. E) Side view of the user showing the torque ($\Gamma_p$) due to the offset between the device and the arm weight.}
    \label{fig:userStudy}
\end{figure}

The aim of this user study was to investigate the SBS performance to prevent arm fatigue and its acceptance by users, and to compare it with both the previously developed GSS (see \cite{rognon2018flyjacket} and Figure \ref{fig:GSS}), and with NS when controlling a simulated fixed-wing drone. The simulated drone  was developed in Unity3D (Unity Technologies, San Francisco, CA, USA) and its physics were based on the eBee (SenseFly, Parrot Group, Paris, France), flying at a constant cruise speed of 12 $\frac{m}{s}$ \cite{cherpillod2019embodied}.

Twelve participants (Ten men and two women, age 32.55 $\pm$ 8.87 years; mean $\pm$ SD; height min = 1.62 m, max = 1.99 m; weight min = 59 kg, max = 86 kg) took part in this experiment. The EPFL Institutional Review Board approved the study and the participants provided written informed consent. 

To study the devices' efficiency to reduce arm fatigue, we recorded the bilateral electromyographic activity (EMG, Desktop DTS, Noraxon, USA) of six shoulder muscles: the anterior (DANT), medial (DMED) and posterior deltoid (DPOS), the infraspinatus (INF), and the upper (TRAPU) and middle trapezius (TRAPM), as well as the biceps (BIC) for every task of the experiment.

After placing the EMG electrodes, the participants performed a Maximal Voluntary Contraction (MVC) test for each muscle. Next, the participants put on the FlyJacket above the electrodes and sat on a stool (Figure~\ref{fig:userStudy}). They also wore virtual reality goggles (Oculus Rift, Facebook, CA, USA) that gave a first person view of the flight and wind sound for more immersion. 
The experiment started with a short training without arm support composed of two tasks. First, the participants had to follow the direction of an arrow positioned in front of them. The arrow was pointing consecutively “right”, “left”, “up”, and “down” twice. The goal of this task, which lasted one minute, was to make the participants perform every flight control movement at least once. The second task was one and a half minutes of free flight in a 3D reconstruction of the EPFL campus. The goal of the training was to allow the participants to feel comfortable with the control of the flight.
After this short training, the participants had to fly three times through 50 waypoints represented by small clouds (Figure~\ref{fig:userStudy} B) once with the SBS, once with the GSS and once with NS. The order of these flights was pseudo-randomized between the participants. The waypoints formed a trajectory in the sky and disappeared when they were reached. The waypoint sequence was randomized, but the number of maneuvers (up/down/right/left) was the same for every task. The flight performance was computed as the Root Mean Square (RMS) of the distance between the drone and each waypoint. The participants had a five-minute break between each task, during which the type of arm support was changed and adapted to their morphology.
At the end of the flight tasks, they filled out a questionnaire about their appreciation of each arm support.

All calculations for the data analysis were computed in MATLAB and R Studio (R Studio Inc., Boston, MA, USA).
The raw EMG data, acquired at 1500Hz were corrected for linear trends. The signals were then low-pass filtered at 400 Hz, high-pass filtered at 50 Hz, rectified, and low-pass filtered at 5 Hz to remove noise using seventh-order Butterworth filters. Eventually, each channel was normalized to its MVC. 

\subsubsection{Results}

\begin{figure*}[tb]
    \centering
    \includegraphics[width=2\columnwidth]{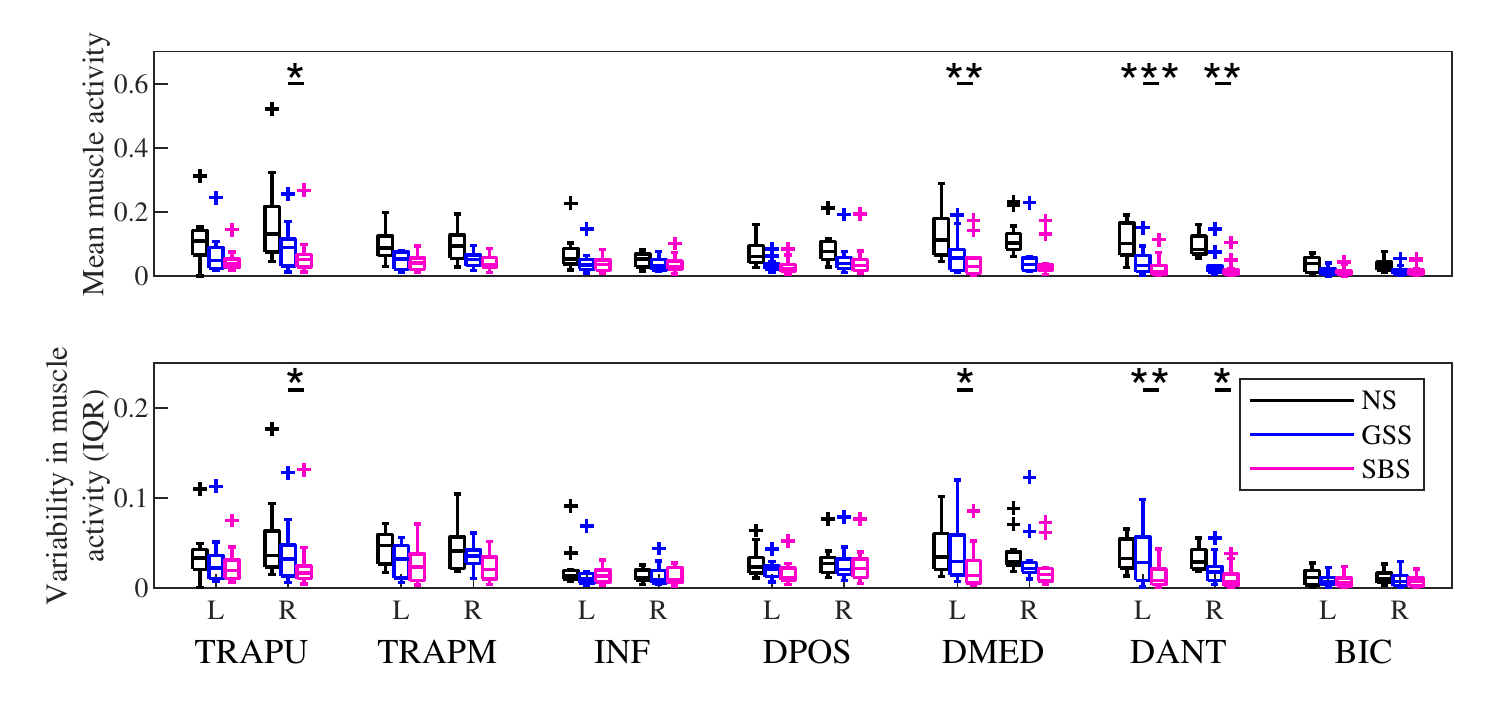}
    \caption{Muscular activities (n=12). A) Mean muscle activity. B) Variability as assessed by the interquartile range (IQR). Significant differences are indicated only between GSS and SBS conditions; *$p<$ 0.05, **$p<$ 0.01, ***$p<$ 0.001}
    \label{fig:muscleActivity}
\end{figure*}

 \begin{table*}[tb]
     \caption{Statistical Significance of the Different Arm Supports on the Mean Muscle Activity (n=12). Post-hoc t-tests were only conducted for the muscles on which the supports showed a significant effect}
     \centering
     \begin{tabular}{@{}llcccccccc@{}}
     \toprule
      &  & \multicolumn{2}{c}{\textbf{ANOVA}} & \multicolumn{2}{c}{\textbf{NS vs GSS}} & \multicolumn{2}{c}{\textbf{NS vs SBS}} & \multicolumn{2}{c}{\textbf{GSS vs SBS}} \\
     \multicolumn{2}{l}{Muscle} & \textit{p} & \textit{$\eta^2$} & \textit{p} & \textit{d} & \textit{p} & \textit{d} & \textit{p} & \textit{d} \\ \midrule
     \multirow{2}{*}{TRAPU} & L & 5.83 $\cdot 10^{-4}$ & 0.63 & 0.031 & 0.83 & 3.84 $\cdot 10^{-3}$ & 1.24 & 0.097 & 0.52 \\
     & R & 5.83 $\cdot 10^{-4}$ & 0.68 & 7.40 $\cdot 10^{-3}$ & 1.06 & 2.25 $\cdot 10^{-3}$ & 1.33 & 0.024 & 0.75 \\
     \multirow{2}{*}{TRAPM} & L & 7.05 $\cdot 10^{-5}$ & 0.72 & 1.18 $\cdot 10^{-3}$ & 1.37 & 8.54 $\cdot 10^{-4}$ & 1.51 & 0.250 & 0.35 \\
      & R & 1.18 $\cdot 10^{-4}$ & 0.76 & 2.76 $\cdot 10^{-3}$ & 1.22 & 4.14 $\cdot 10^{-4}$ & 1.65 & 0.178 & 0.42 \\
     \multirow{2}{*}{INF} & L & 0.017 & 0.67 & 0.010 & 1.08 & 0.055 & 0.73 & 0.492 & 0.21 \\
      & R & 0.017 & 0.52 & 0.019 & 0.97 & 0.085 & 0.66 & 0.745 & -0.10 \\
     \multirow{2}{*}{DPOS} & L & 2.60 $\cdot 10^{-4}$ & 0.70 & 1.18 $\cdot 10^{-3}$ & 1.45 & 1.18 $\cdot 10^{-3}$ & 1.44 & 0.318 & 0.30 \\
     & R & 9.77 $\cdot 10^{-5}$ & 0.72 & 1.60 $\cdot 10^{-3}$ & 1.32 & 7.47 $\cdot 10^{-4}$ & 1.53 & 0.342 & 0.29 \\
     \multirow{2}{*}{DMED} & L & 1.33 $\cdot 10^{-4}$ & 0.77 & 1.02 $\cdot 10^{-3}$ & 1.40 & 3.44 $\cdot 10^{-4}$ & 1.68 & 4.98 $\cdot 10^{-3}$ & 1.01 \\
      & R & 2.60 $\cdot 10^{-4}$ & 0.71 & 9.21 $\cdot 10^{-4}$ & 1.49 & 2.60 $\cdot 10^{-3}$ & 1.24 & 0.420 & 0.24 \\
     \multirow{2}{*}{DANT} & L & 6.52 $\cdot 10^{-5}$ & 0.83 & 2.58 $\cdot 10^{-4}$ & 1.66 & 7.50 $\cdot 10^{-5}$ & 2.00 & 5.98 $\cdot 10^{-4}$ & 1.37 \\
      & R & 7.46 $\cdot 10^{-7}$ & 0.92 & 3.31 $\cdot 10^{-5}$ & 2.09 & 1.09 $\cdot 10^{-6}$ & 3.10 & 9.29 $\cdot 10^{-3}$ & 0.91 \\
     \multirow{2}{*}{BIC} & L & 1.61 $\cdot 10^{-3}$ & 0.60 & 5.76 $\cdot 10^{-3}$ & 1.15 & 5.76 $\cdot 10^{-3}$ & 1.17 & 0.358 & 0.28 \\
      & R & 6.52 $\cdot 10^{-5}$ & 0.75 & 4.98 $\cdot 10^{-4}$ & 1.55 & 4.98 $\cdot 10^{-4}$ & 1.61 & 0.898 & -0.04 \\ \bottomrule
     \end{tabular}
     \label{tab:statsMeans}
 \end{table*}

\begin{table*}[tb]
    \caption{Statistical Significance of the Different Arm Supports on the Variability (n=12). Post-hoc t-tests were only conducted for the muscles on which the supports showed a significant effect}
    \centering
\begin{tabular}{@{}llllllllll@{}}
\toprule
 &  & \textbf{ANOVA} &  & \textbf{NS vs GSS} &  & \textbf{NS vs SBS} &  & \textbf{GSS vs SBS} &  \\ 
\multicolumn{2}{l}{Muscle} & \textit{p} & $\eta^2$ & \textit{p} & \textit{d} & \textit{p} & \textit{d} & \textit{p} & \textit{d} \\\midrule
TRAPU & L & 0.165 & 0.27 & - & - & - & - & - & - \\
 & R & 6.38 $\cdot 10^{-4}$ & 0.69 & 0.014 & 0.95 & 2.00 $\cdot 10^{-3}$ & 1.39 & 0.028 & 0.73 \\
TRAPM & L & 7.97 $\cdot 10^{-4}$ & 0.69 & 0.003 & 1.20 & 3.00 $\cdot 10^{-3}$ & 1.29 & 0.226 & 0.37 \\
 & R & 9.53 $\cdot 10^{-3}$ & 0.58 & 0.14 & 0.55 & 9.00 $\cdot 10^{-3}$ & 1.09 & 0.14 & 0.58 \\
INF & L & 0.266 & 0.52 & - & - & - & - & - & - \\
 & R & 0.843 & 0.04 & - & - & - & - & - & - \\
DPOS & L & 0.024 & 0.42 & 0.065 & 0.71 & 0.052 & 0.81 & 0.126 & 0.48 \\
 & R & 0.178 & 0.44 & - & - & - & - & - & - \\
DMED & L & 0.019 & 0.73 & 0.878 & -0.05 & 3.00 $\cdot 10^{-3}$ & 1.29 & 0.015 & 0.94 \\
 & R & 0.082 & 0.33 & - & - & - & - & - & - \\
DANT & L & 3.71 $\cdot 10^{-3}$ & 0.80 & 0.853 & 0.06 & 3.71 $\cdot 10^{-3}$ & 1.80 & 0.009 & 1.04 \\
 & R & 6.38 $\cdot 10^{-4}$ & 0.74 & 0.018 & 0.80 & 1.00 $\cdot 10^{-3}$ & 1.50 & 0.018 & 0.91 \\
BIC & L & 0.035 & 0.38 & 0.103 & 0.63 & 0.074 & 0.75 & 0.422 & 0.24 \\
 & R & 0.066 & 0.35 & - & - & - & - & - & - \\ \bottomrule
\end{tabular}
    \label{tab:statsIQR}
\end{table*}

\begin{figure}[tb]
    \centering
    \includegraphics[width=\columnwidth]{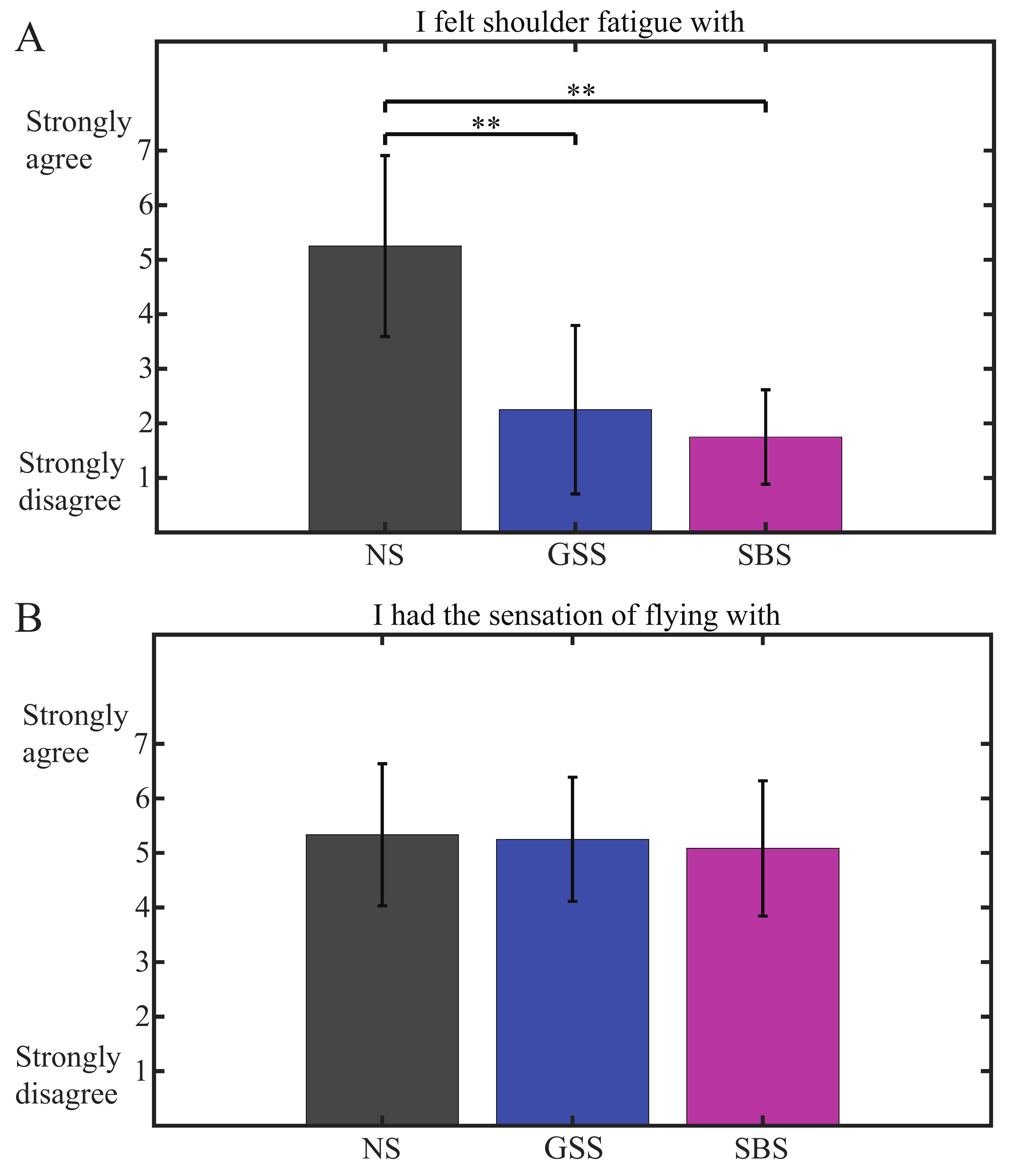}
    \caption{Questionnaire results (n=12). A) for the question "I felt shoulder fatigue". B) for the question "I had the sensation of flying". (**) denotes $p < 0.01$.}
    \label{fig:qresults}
\end{figure}

There were no significant differences in the flight performance between the three flight conditions with a RMS error of 5.31 m $\pm$ 8.26 (mean (m) $\pm$ STD) for the SBS, 3.62 m $\pm$ 4.04 for the GSS, and 3.75 m $\pm$ 4.00 for NS, which means that the SBS does not affect the users' performance.

Figure~\ref{fig:muscleActivity} A shows the mean muscle activity for the six recorded shoulder muscles, and for the biceps. Figure \ref{fig:muscleActivity} B shows the variability in the muscle activity between the 25$\%$ and 75$\%$ quartiles. The variability displays the mechanical stability of the support whereas the mean muscle activity shows the efficiency of the support in assisting the muscles. No significant difference in the ANOVA results of the variability can mean either that support has no effect or that these muscles were not active during flight. Repeated measure analyses of variance (ANOVAs) revealed a significant effect of support for all muscles on the mean activity (see Table \ref{tab:statsMeans} for the statistical analysis values, $p$-values were corrected for multiple comparisons using the Benjamini-Hochberg procedure). The statistical differences between each pairs of supports (SBS, GSS, and NS) was tested with t-tests. Using any arm support significantly reduces the muscle activity when comparing with NS. The effect sizes, assessed with Cohen's $d$, show a large effect for most of the muscles with values higher than 0.8, strengthening the importance of the support.
A similar analysis also showed a significant effect of the supports on the variability of the activity for eight muscles (see Table \ref{tab:statsIQR}). 

Statistically significant differences between the two arm supports (SBS and GSS) were observed in both the mean muscle activity and their variability for the right TRAPU, the left DMED, and both left and right DANT . 
As a large part of users fly with the elbow bent instead of straight --  9/12 participants of the user study adopting this flight style -- they probably use their DANT to support the torque produced by their forearm in this position. The SBS seems to be providing more support for the shoulder rotation than the GSS, consequently reducing the muscle activity and the variability of the DANT.  
A statistically significant effect was also observed for the right TRAPU, and the left TRAPU t-test shows a $p$-value relatively close to significance ($p$ = 0.097). The TRAPU is playing an active role in the stabilization of the shoulder both in abduction and in rotation. These results reinforce our assumption of the SBS being a more stable device for the shoulder rotation than the GSS.
Results for the DMED are surprising as only the left muscle showed a difference in muscle activity and variability between the two arm supports. Indeed, we expected to have a symmetrical effect of the supports. As only one participant was left-handed, we cannot draw conclusion on the possibility of stronger muscles on one side of the body due to handedness. Another explanation could be a difference in the mechanical characteristics between the left and the right arm support (which could exist in either the SBS or the GSS). However, this difference in muscle activity between body sides is present only for the DMED.

Wilcoxon Rank Sum tests performed on the questionnaire data, showed that the participants felt significantly less fatigue when flying with the arm supported by the SBS ($p=1.2 \cdot 10^{-4}$) or with the GSS ($p=8.6 \cdot 10^{-4}$) (which corroborates previous results \cite{rognon2018flyjacket}) than with NS (Figure~\ref{fig:qresults} A). There are no significant differences in subjective shoulder fatigue between the two arm supports. Both arm supports did not prevent users to have enjoyable sensations, as the sensation of flying was rated the same for the three flight conditions (Figure~\ref{fig:qresults} B). The participants felt equally confident with both arm supports and did not feel more constrained with one arm support comparing to the other. Some participants reported that they felt more confident flying with an arm support as they did not have to reflect about what to do with their arms.
\section{Discussion}
The novel arm support presented in this article addresses the challenges of compensating the arms weight along the ROM used to fly a drone using body movements, while remaining portable and adapting to multiple body morphologies. Indeed, this new arm support is adaptable to the body size of more than 97$\%$ of the population. With a mean torque error for the population range extremities of -0.005 $\frac{\text{Nm}}{\text{kg}}$ for the 1PF and -0.002 $\frac{\text{Nm}}{\text{kg}}$ for the 99PM over the ROM, it enables a better torque compensation along a larger ROM than the arm support developed in our previous work \cite{rognon2018flyjacket}. Indeed, the previous arm support has a torque error 130 times higher for the 1PF and 30 times for the 99PM, while covering less than half of the ROM. To our knowledge, this novel arm support has a better overall performance (when combining the torque error, ROM, body type fit and weight of the device) than currently developed arm supports. The parasitic force is also substantially lower across the whole ROM than for the previous arm support, on average 117 times smaller for the 1$^{st}$ percentile female and 21 times for the 99$^{th}$ percentile male, allowing a more comfortable device for the user. No parasitic force measurements have been found in literature. The mechanical characterization of the built device validated the feasibility of a torque compensation up to 18 Nm ensuring arm masses going up to 5.87 kg for an upper arm length of 0.312 m to be fully compensated. The results of the user study confirmed the usefulness of an arm support during flight as the muscle activity of the shoulder was significantly reduced (on average 58\% than without an arm support) and the participants felt less shoulder fatigue. In addition, the participants reported to have the same flight sensation with arm support as without. In comparison to the previously developed arm support, the results of the muscle activity recorded on the shoulder suggest that the SBS gives more support to the shoulder with a mean muscle activity reduction of 34\% for the deltoids muscles.

The aim of the user study being to compare the performance of the SBS in comparison to the previously developed arm support (GSS), we tested the performance with the benchmark task and command strategy. The current flight strategy involves torso movements for controlling the drone, while the arms are passively spread out to improve performance. Even for a task involving only passively the arms, reduction in some shoulder muscle activities could be observed, and neither the flight performance nor the user acceptance were diminished despite the augmentation of the support complexity. We foresee that the accurate weight compensation offered by the SBS along a larger ROM will become even more significant in tasks involving actively arms movements to control the drone. For example, future work will investigate the implementation of a machine learning algorithm that learns the natural gestures of the full upper body of the user and adapts the control strategy accordingly to each individual. In this case, due to the large variation between individuals and the active use of the arms, the ROM of gestures and positions where the gravity needs to be compensated requires to be larger than what the GSS can provide.
    
Some minor improvements can be done to the SBS. As the thread of the screw is 1.5mm, screwing the grounding part until the gravity is compensated -- on average 40 mm for the participants of the user study -- takes time. As such a precise positioning of the grounding part is not required in our systems, this screw could be replaced by one with a larger thread. In the case where a precise positioning of the screw is however required, the tuning of the torque could be motorized. Another issue is that, due to the placement of the knob on the back of the user, the adjustment for the torque compensation cannot be done by the users themselves. A motorized torque with a digital interface would also address this issue.

In a future work, haptic feedback could be transmitted through the arm support in order to render flight sensations (such as air lift or drag) or guidance to the FlyJacket user. Actuators could be added in this arm support to render these forces. However, a special attention needs to be given to keep the device portable.



\section*{acknowledgment}
This work was supported by the Swiss National Science Foundation (SNSF) through the National Centre of Competence in Research Robotics (NCCR Robotics) and through the FLAG-ERA project RoboCom++.


\bibliographystyle{elsarticle-num-names}
\bibliography{main}

\appendix       
\section*{Appendix A: Force and Torque Calculations}
\label{append}

The torque required to compensate the arm weight can be computed depending on the arm position.

\begin{figure}[t] 
    \centering
    \includegraphics[width=\columnwidth]{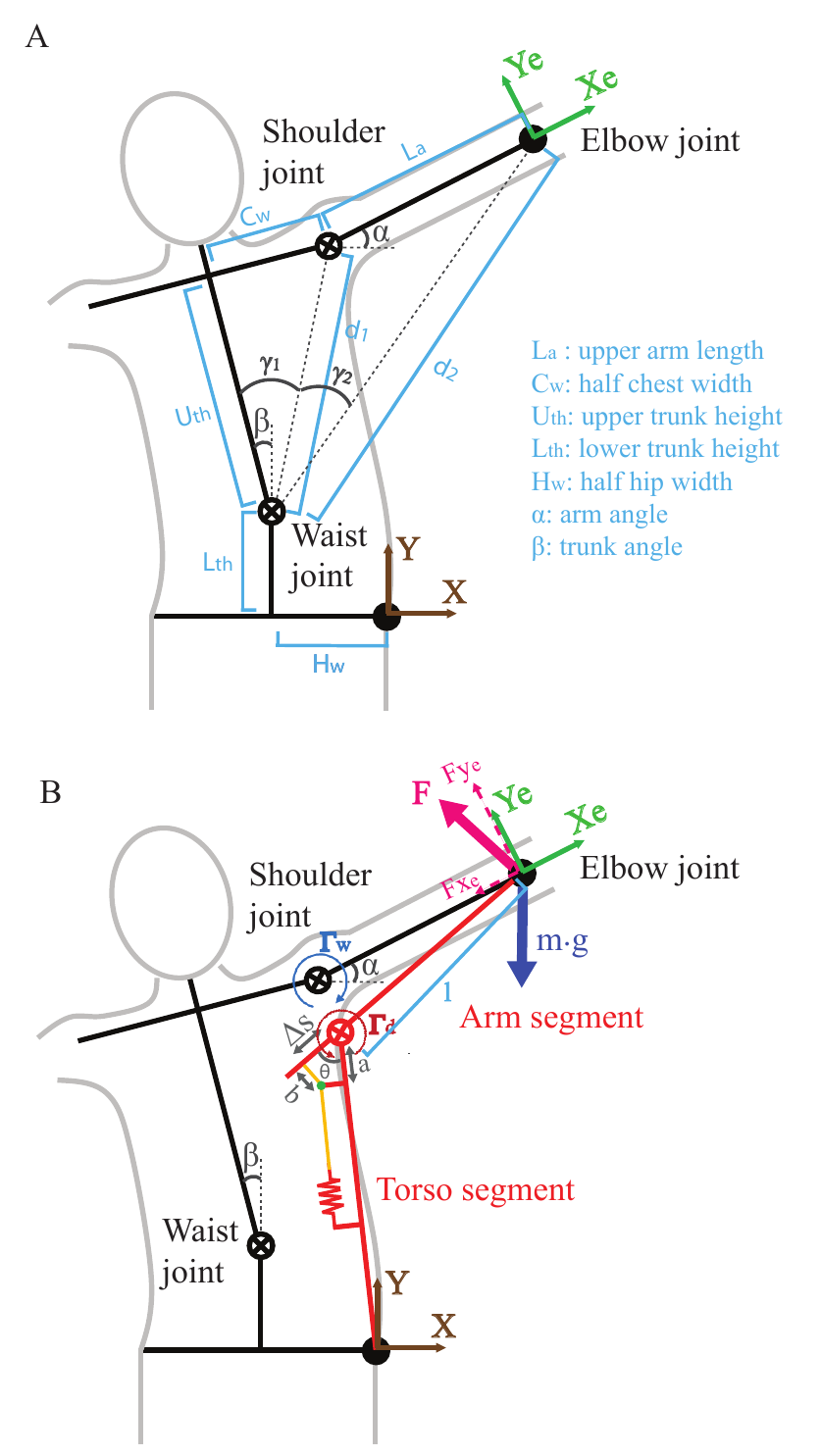}
    \caption{Schema of the upper body.  A) Body dimensions. B) Schema of the device, and forces and torques induced.}
\label{fig:ModelAppend}
\end{figure}

At first, the output positions of the elbow joint can be calculated using the parameters described in Figure~\ref{fig:ModelAppend} A.
\begin{equation}
    d_1 = \sqrt{u_{th}^{2} + c_{w}^{2}}
\end{equation}
\begin{equation}
    \gamma_1 = \arctan(\frac{c_{w}}{u_{th}})
\end{equation}
\begin{equation}
    d_2 = \sqrt{L_{a}^{2} + d_1^{2} - 2 \cdot L_{a} \cdot d_1 \cdot \cos(\frac{\pi}{2} + \gamma_1 + \alpha)}
\end{equation}

\begin{equation}
    \gamma_2 = \arccos{(\frac{d_1^{2} + d_2^{2} - L_{a}^{2}}{2 \cdot d_1 \cdot d_2})}
\end{equation}

\begin{equation}
    x= d_2^{2} \cdot \sin(\alpha + \gamma_1 + \gamma_2) - H_w
    \label{eq:xROM}
\end{equation}
\begin{equation}
    y= d_2^{2} \cdot \cos(\alpha + \gamma_1 + \gamma_2) + L_{th}
    \label{eq:yROM}
\end{equation}

Upon the output positions, the gravity force values can be calculated. In the current model, two different coordinate systems are used: The XY coordinate system (brown in Figure~\ref{fig:ModelAppend}) for the positions and the $XeYe$ coordinate system (green in Figure~\ref{fig:ModelAppend}) for the force projections. This allows to decompose the force generated by the mechanism in two components: the gravity compensation contribution on $Ye$ and the parasitic forces in the arm on $Xe$ (see Figure~\ref{fig:ModelAppend} B). The forces on $Xe$ are parasitic and unwanted, even if the gravity also has a component in that direction. Unlike the $Ye$-projected force that counterbalance the arm weight, the $Xe$-projected forces act on the interface with the skin. This will lead the interface to create shear stress on the skin and therefore discomfort for the subject.

Both forces can be computed as in Equation \ref{eq:fvref} and \ref{eq:furef}.
\begin{equation}
    F_{Ye,ref} = \cos{(\alpha + \beta)} \cdot m \cdot g
    \label{eq:fvref}
\end{equation}

\begin{equation}
    F_{Xe,ref} = \sin{(\alpha + \beta)} \cdot m \cdot g
    \label{eq:furef}
\end{equation}

\noindent with $m$ the mass of the arm (see Table \ref{tab:size}) and $g$ the gravity. The error of the arm weight compensation can be calculated as in Equation~\ref{eq:ferrv}.  
\begin{equation}
    F_{errY_e} = F_{Y_e} - F_{Y_e,ref}
    \label{eq:ferrv}
\end{equation}
This force error has to be minimize over the whole domain, which will characterize the performance ($perf$) of the device.

\begin{equation}
    p_e=  \min \Vert\sum_{\alpha, \beta \in ROM} F_{errY_e}(\alpha, \beta, k, l_s, l_o)^2\Vert
    \label{eq:min}
\end{equation}
\noindent with $k$ the spring constant, $l_s$ the spring length, and $l_o$ the initial spring length

The spring is making the gravity compensation mechanism and is ruled by Equation~\ref{eq:spring}.
\begin{equation}
    F_s = k \cdot \Delta x + F_{0}
    \label{eq:spring}
\end{equation}
With $k$ the spring constant ($\frac{N}{m}$), $\Delta x$ the deformation of the spring ($m$), $F_{0}$ the force at initial length ($N$).

In a gravity compensation device, the torque induced by the arm weight $\Gamma_w$ has to be compensated at the shoulder by the torque produced by the device $\Gamma_d$.
\begin{equation}
    \sum^{\Gamma} = \Gamma_d - \Gamma_w 
\end{equation}

The device's torque is:
\begin{equation}
    \Gamma_d = a \cdot F_s \cdot \sqrt{1 - (\frac{a^2 + b^2 - \Delta s^2}{2 \cdot a \cdot b})^2} 
    \label{eq:shoultorque}
\end{equation}

with b being:
\begin{equation}
    b = \sqrt{a^2 + \Delta s^2 - 2 \cdot a \cdot \Delta s \cdot \cos{\theta}}
    \label{eq:xdef}
\end{equation}

After combining Equations~\ref{eq:shoultorque} and \ref{eq:xdef}, we obtain:
\begin{equation}
    \Gamma_d = a \cdot F_s \cdot \frac{\Delta s \cdot \sin{\theta}}{b}
\end{equation}

The torque induced by the arm weight is:
\begin{equation}
    \Gamma_w = l \cdot m \cdot g  \cdot \sin{\theta} 
\end{equation}

As the equilibrium is required, Equation~\ref{eq:equality} must be satisfied:
\begin{equation}
    a \cdot F_s \cdot \frac{\Delta s}{b} = l \cdot m \cdot g
    \label{eq:equality}
\end{equation}

Therefore, taking into account Equation~\ref{eq:spring}, we get a gravity compensation when:
\begin{equation}
    a \cdot \Delta s \cdot \frac{k \cdot (b-l_0) + F_0}{l \cdot b} = m \cdot g
\end{equation}

$b$ being dependant on $\theta$, perfect gravity compensation does not occur. To perform perfect gravity compensation, we can simulate a zero free-length spring by adding a pretension $b_0$ on the spring:
\begin{equation}
    k \cdot (b - l_0 + b_0) + F_0 = k \cdot b \xleftrightarrow{} b_0 = l_0 - \frac{F_0}{k}
\end{equation}

In our device, the distance $a$ is constant, leaving us with one optimization parameter $\Delta s$. In that case, the analytical solution for the torque is:
\begin{equation}
    \Gamma =m \cdot g \cdot l = a \cdot \Delta s \cdot k
    \label{eq:tocalculatetorque}
\end{equation}

\end{document}